\documentclass[10pt, twocolumn, letterpaper]{article}
\usepackage[a4paper, left=20mm, right=20mm, bottom=30mm, top=15mm]{geometry}
\usepackage{cite}
\usepackage{graphicx}
\usepackage{indentfirst}
\usepackage{amsmath}
\usepackage{booktabs}
\usepackage{blindtext}
\usepackage{caption2}
\usepackage{multirow}
\usepackage{float}

\begin{document}
	\date{}
	
	\title{Multi-Level Recurrent Residual Networks for Action Recognition}
	\author{Zhenxing ZHENG$^{1, 2}$\\
		\and
		Gaoyun AN$^{1, 2}$\\
		\and
		Qiuqi RUAN$^{1, 2}$\\
		\and
		$^1$Institute of Information Science, Beijing Jiaotong University, Beijing 100044, China\\
		\and
		$^2$Beijing Key Laboratory of Advanced Information Science\\ and Network Technology, Beijing 100044, China\\
		\and
		\{zhxzheng, gyan, qqruan\}@bjtu.edu.cn}
	
	\maketitle
	
	\begin{abstract}
		\textit{ Most existing Convolutional Neural Networks(CNNs) used for action recognition are either difficult to optimize or underuse crucial temporal information. Inspired by the fact that the recurrent model consistently makes breakthroughs in the task related to sequence, we propose a novel Multi-Level Recurrent Residual Networks(MRRN) which incorporates three recognition streams. Each stream consists of a Residual Networks(ResNets) and a recurrent model. The proposed model captures spatiotemporal information by employing both alternative ResNets to learn spatial representations from static frames and stacked Simple Recurrent Units(SRUs) to model temporal dynamics. Three distinct-level streams learned low-, mid-, high-level representations independently are fused by computing a weighted average of their softmax scores to obtain the complementary representations of the video. Unlike previous models which boost performance at the cost of time complexity and space complexity, our models have a lower complexity by employing shortcut connection and are trained end-to-end with greater efficiency. MRRN displays significant performance improvements compared to CNN-RNN framework baselines and obtains comparable performance with the state-of-the-art, achieving 51.3\% on HMDB-51 dataset and 81.9\% on UCF-101 dataset although no additional data.}
	\end{abstract}

	\section{Introduction}
	With the development of deep learning and the improvement in computer hardware, action recognition attracts growing attention in the research community\cite{Karpathy2014Large,Simonyan2014Two, Tran2014Learning}. There are many potential applications of action recognition, like video caption, abnormal event detection, intelligent monitoring, and auto drive. However, action recognition remains a fundamental challenge in computer vision, since it is affected by rapid movement, illumination variation, occlusion and viewpoint variation largely.
	
	\begin{figure}[tbp]
		\centering
		\includegraphics[width=8cm]{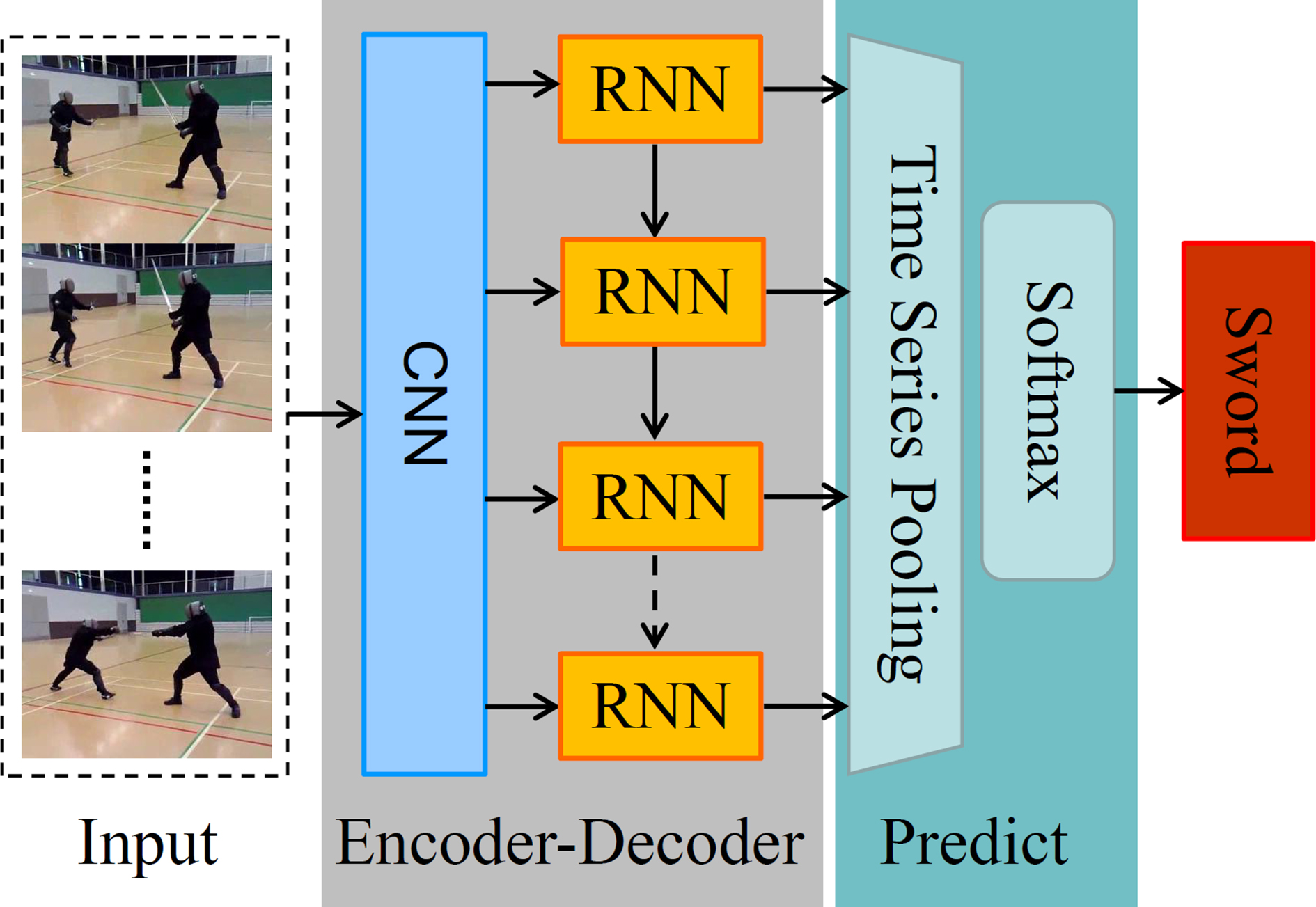}
		\caption{\textbf{CNN-RNN framework.} \small{A generic model, CNN with Recurrent Neural Networks(RNN) for action recognition. CNN is applied to encode the appearance of video frames into a group of fixed length vectors, frame by frame, that lately is decoded through RNN to learn video representation spatially and temporally. Then, time series pooling is used over outputs at all time-steps followed by a softmax layer to predict activity.}}\label{sketchmap}
	\end{figure}
	
	In the early study, researchers mainly focus their attention on the visual representation of the static image which contains no motion information. The survey on still image\cite{Guo2014A} shows high-level cues, including visual components(e.g., human body\cite{Li2012What}, object\cite{Alexe2010What}, scene\cite{Li2007What}) appeared in the image, and low-level features, including scale invariant feature transform(SIFT)\cite{Yao2010Grouplet} and Histogram of oriented gradients(HOGs)\cite{Dalal2005Histograms}, are pursued by researchers for the sake of characterizing actions. Given a video sequence, it is natural to decompose into spatial and temporal components. Compared with the traditional still image based action recognition, videos provide additional temporal information for distinguishing behaviors. To utilize temporal information, several hand-crafted features such as Space Time Interest Points(STIP)\cite{Laptev2005On}, dense trajectories with motion boundary histograms(MBH) Descriptors\cite{wang:2011:inria-00583818:1}, SURF descriptors with the dense optical flow\cite{Wang2014Action} are widely used in action field on account of needing no algorithm to detect human body. They usually detect discriminative regions for action analysis first and subsequently encode these local features into one vector as the representation of the image which is used to learn classifiers for action recognition. Among these local features, improved trajectories\cite{Wang2014Action} achieves the best performance on several challenging benchmarks(e.g., HMDB-51\cite{Kuehne11}, Hollywood2\cite{marszalek09}) enormously. 
	
	Recently, Convolutional Neural Networks(CNNs)\cite{Lecun1998Gradient} has been demonstrated as an effective way to automatically learn substantial discriminative visual representations and make significant breakthroughs in image classification, especially deep CNNs\cite{Krizhevsky2012ImageNet, Simonyan2014Very}. Inspired by this, many works begin to generalize deep learning methods to the field of action recognition. To learn stronger spatiotemporal representations, 3D convolutional networks\cite{Tran2014Learning}, Trajectory-Pooled Deep-Convolutional Descriptors\cite{Wang2015Action}, and LRCN\cite{DBLP:journals/corr/DonahueHGRVSD14} learn spatiotemporal representations directly. Unlike these, convolutional Two-Stream Networks\cite{Feichtenhofer2016Convolutional}, Multi-region two-stream R-CNN\cite{Peng2016Multi} and cLSTM\cite{Srivastava2015Unsupervised} use two networks to learn spatial and temporal representations independently and then fuse them. The learned temporal representations explicitly from optical flow maps by Two-Stream Networks turn out to be better than C3D relied on motion-sensitive convolutional kernels.
	
\begin{figure*}[tbp]
		\centering
		\includegraphics[width=16cm, height=10.5cm]{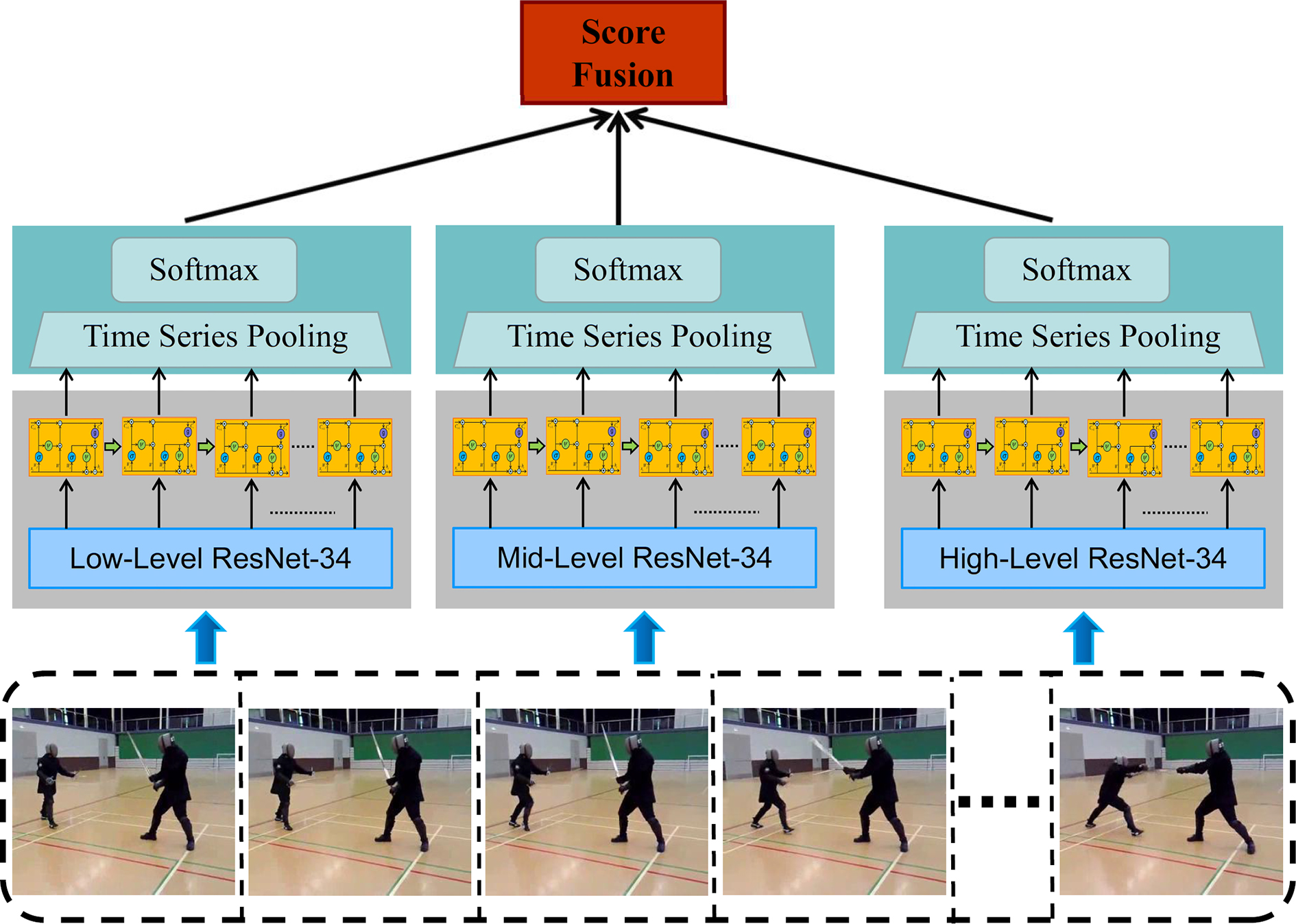}
		\caption{\textbf{Pipline of MRRN.} \small{Our network consists of three independent models. We capitalize on different-level RRN to produce different-level video representations simultaneously. Then these representations are fused by computing a weighted average of their softmax scores to obtain the complementary representations for action recognition.}}\label{piplines}
	\end{figure*}

	Despite good performance, these methods are either computationally intensive or insufficient in space domain. To deal with these, we propose a novel Multi-Level Recurrent Residual Networks(MRRN) model for action recognition, as illustrated in fig.\ref{piplines}. In the proposed model, ResNets is applied to encode the appearance of video frames into fixed length vectors that lately are decoded through stacked SRUs to produce video representations. Employing identity shortcut connections in MRRN is parameter-effective, which lowers space complexity and time complexity by a large margin. Then, time series pooling is used over the entire outputs at all time-steps and predict the probability score by softmax layer. Considering the fact that higher activation values of different depths in the network gather around different parts of the image, for example, mid-level representations focus on legs while high-level representations focus on the whole body, we develope three different-level models to produce complementary representations and make final predictions. In MRRN, three separate sub-models are called low-, mid-, high-level Recurrent Residual Networks(RRN) respectively. 
	
	The contributions of this paper are shown as follows: First, we analyze the effect of diverse hyper-parameter settings qualitatively to illustrate the general tendency of performance. To lower the space complexity and time complexity, we propose to use identity shortcut connections in the proposed model. Additionally, we experiment with low-, mid-, high-level representations of the video in various time pooling manners, experimentally demonstrating how well different-level features contribute to action recognition. Based on the complementary relation between the whole and the part, our proposed architecture consists of three separate recognition streams captured different-level information to learn effective representations for action recognition. Finally, a series of experiments were carried out on two standard video actions benchmarks of HMDB-51 and UCF-101 dataset. Experimental results show MRRN displays significant performance improvements compared to CNN-RNN framework baselines and obtains comparable performance with the state-of-the-art.
	
	The rest of the paper is organized as follows. In section 2, we review various state-of-the-art methods addressing partly challenging problems in action recognition. The method we proposed in this paper is described in section 3. Implement details are introduced in section 4. We make analyses on the experimental results in section 5. Finally, we draw conclusions.

	\section{Related Work}
	\textbf{Hand-crafted features.} Local features are of popularity in image classification, which characterizes images through descriptors such as SIFT, HOGs, and SURF. Image-based action recognition concentrates on identifying actions appeared in the static image. Exemplarlet\cite{Li2012What} comprises abundant visual information (e.g., pose) within the body for discerning human actions. For this purpose, what need to do is manually selecting and segmenting bounding box in images. Objectness method\cite{Alexe2010What} quantifies the probability of the bounding box that contains the object of any class, so multiple candidates relevant to actions (e.g., bike, basketball, laptop) can be found from the cluttered background. The integrative model\cite{Li2007What} integrates scene and object categorizations to discriminate events occurred in images. Nevertheless, low-level features usually can't work well alone due to the cluttered background and crucial temporal information discarded by these methods. In order to address above problems, previous research directly extends the recognition algorithms based on the static image to learn spatiotemporal representations. Traditional video-based action recognition\cite{Poppe2010Poppe} is described by a collection of local descriptors. For example, extended from the Harris corner detector\cite{Harris1988A}, Harris3D detector\cite{Laptev2005On} was proposed to encode the region of the interest(ROI). RMM\cite{Zhao2017Region} encodes the layout of hybrid features for action discrimination. Unlike this, Dense trajectories\cite{wang:2011:inria-00583818:1} tracks dense points that are sampled in each frame depending on the dense optical flow field. This method is shown to be high capacity for video representations. Based on this work, Improved dense trajectories\cite{Wang2014Action} takes camera motion into account and removes the trajectories which are consistent with the camera, achieving state-of-the-art results. Whereas, extracting hand-crafted features along the trajectory have higher computational complexity. 
	
	\textbf{Deep-learning features.}
	Convolutional Networks has shown it can extract deep spatial features for image classification. Encouraged by the impressive performance, many works make extensive attempts at using deep-learning methods for action recognition. The prior attempt is the strategies\cite{Karpathy2014Large} of fusing information across the time domain such as late fusion, early fusion or slow fusion so that high layers acquire plentiful spatiotemporal information. 3D Convolutional Neural Networks(3D CNN) directly captures spatiotemporal information from multiple adjacent frames by replacing the $2\times 2$ convolutional kernel with $3\times3$. Beyond that, an alternative way of associating RGB appearance with motion is Two-Stream Convolutional Networks\cite{Simonyan2014Two}. This model contains two identical networks where one net for RGB images and another for stacked optical flow images. Two networks are trained independently and then combined by using average or linear SVM methods to fuse softmax scores of each. Recently, considering the Long Short-Term Memory(LSTM) playing a vital role in the task related to sequence, CNN with LSTM\cite{DBLP:journals/corr/SharmaKS15} becomes an active research topic in action recognition. Specifically, the feature sequences are extracted by CNN from video frames and then passed into LSTM step by step for action recognition, as illustrated in fig.\ref{sketchmap}.
	
	Motivated by above analysis, we proposed a novel model, Multi-Level Recurrent Residual Networks(MRRN), which integrates three distinct-level ResNet-34 with stacked SRUs and use complementary representations to classify actions. The details will be introduced in the next section.

	\section{Method}
	In the proposed MRRN model, different-level ResNet-34 which was pre-trained on the challenging image classification dataset called ImageNet\cite{Deng2009ImageNet} is used to extract different-level representations of the video sequence. To form the compositional representation in time domain, stacked SRUs\cite{lei2017sru} processes hierarchical features further followed by a pooling layer and a softmax layer for predicting the activity. In the end, three different-level models are combined by late fusion.
	
	\subsection{Residual Networks}
	Deep architectures\cite{Szegedy2015Going, Simonyan2014Very} play an important role in ImageNet Large-Scale Visual Recognition Challenge2014 (ILSVRC2014)\cite{Russakovsky2014ImageNet} and reveal that adding network depth is of importance for improving performance and enriches hierarchical features. On the other hand, the deep system is difficult to optimize on account of notorious vanishing and exploding gradients\cite{Glorot2010Understanding} which impede convergence. In addition, accuracy plateaus even degrades.
	
	Recently, several novel networks with more than 100 layers are trained well by adding shortcut connection\cite{He2015Deep} or dense connection\cite{Huang2016Densely} to mitigate overfitting. These deep networks set a series of records of highly challenging object recognition and classification benchmarks. Note that attention maps\cite{Zagoruyko2016Paying} show higher activation values of different depths in networks gather around different parts of the image, and we subsequently investigate how well diverse representations work for taking activity predictions.
	
	The Residual Networks(ResNets) constructs an extremely deep network by formulating a desired residual mapping as $ \Phi (x) $ and fitting another mapping of $ F(x)=\Phi (x)-x $ for each stacked layers called the building block. Each block contains two layers or three layers according to the requirement of depth, where two layers are $3\times3$, $3\times3$ convolutional kernels and three layers are $1\times1$, $3\times3$ and $1\times1$ convolutional kernels. Between weight layers, rectification (ReLU) activation function\cite{Glorot2012Deep} is used to introduce nonlinearity. Employing $1\times1$ convolutional kernels can reduce time complexity while retaining similar space complexity. The degrade and optimization problem with the increase of the depth are addressed by learning residual functions with reference to the layer inputs. In general, ResNets consists of multiple residual blocks and perform short connection (identity mapping in this case) from the bottom to the top in each block, where information flows from shallow to deep. 
	
	In this paper, we use ResNet-34 to build different level representations. We choose the output activations of last three groups of residual blocks as low-, mid, high-level representation depicted as $A_{l}, A_{m}, A_{h}$ and name corresponding ResNet-34 with different layers low-, mid-, high-level ResNet-34 respectively. Then these appearance representations are pushed into SRUs to obtain spatiotemporal video representations.

	To represent the structure of ResNet-34, we use shorthand notations expressed as follows: Conv2d, BatchNorm2d, ReLU, MaxPool2d, First Sequential BasicBlock, Second Sequential BasicBlock, Third Sequential BasicBlock, Fourth Sequential BasicBlock, AvePool2d and Linear, where conv2d is 2D convolution, BatchNorm2d is 2D batchnormalization, ReLU is rectified linear units, MaxPool2d is 2D max pooling, Sequential BasicBlock is a group of building blocks, AvePool2d is 2D average pooling and Linear is fully connection. The last three Sequential BasicBlock and its corresponding output activation tensor $A\in  R^{C\times H\times W}$ can be represented as : 
	\begin{equation}\label{representation}
	A_{level} = [A_{l1}, A_{l2}, ..., A_{lN}]\qquad A_{l,i}\in R^{C}
	\end{equation}
	where $level\in [low, mid, high]$ and $N=H\times W$. Specifically, $A_{l}\in  R^{128\times 28\times 28}$, $A_{m}\in  R^{256\times 14\times 14}$, $A_{h}\in  R^{512\times 7\times 7}$. We then average these activations $A\in  R^{C\times H\times W}$ and result in descriptors $x_{feature}\in  R^{C}$ of video frames. 
	\subsection{Recurrent Model}
	\begin{figure}[htbp]
		\centering
		\includegraphics[width=8cm]{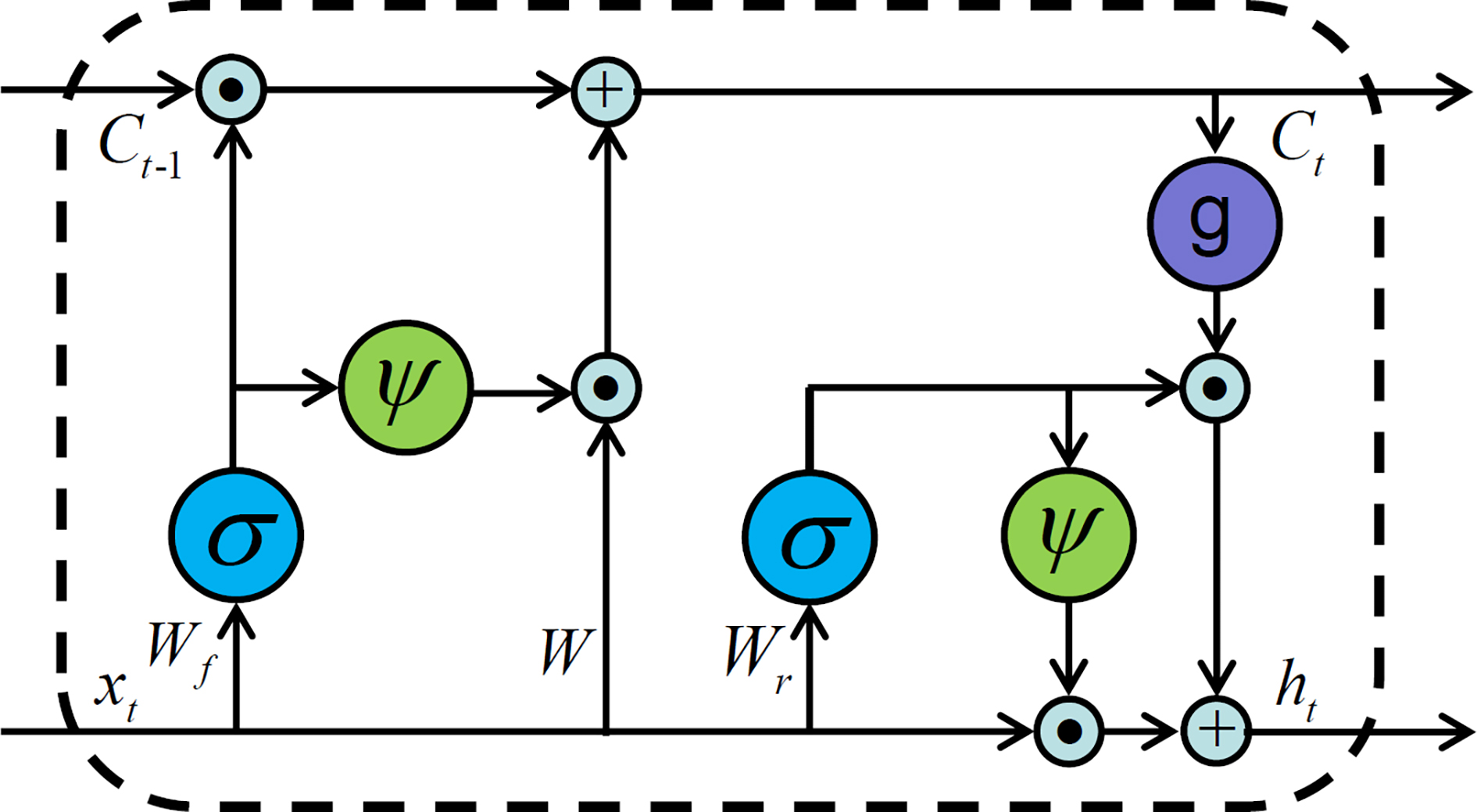}
		\caption{\textbf{SRU Architecture.} \small{$ \psi $ represents the operation of $ 1-input $, $ g $ is the hyperbolic tangent function, $ \sigma $ is the sigmoid function. $ \odot   $ and $ \oplus  $ are element-wise multiplication and addition respectively. We did not show the bias for simplication.}}\label{sru}
	\end{figure}
	RNNs, especially LSTM\cite{Graves1997Long} are widely used in machine translation, image caption and speech recognition, achieving desirable results. The update of gate states in the recursion, however, depends on the previous hidden states $h_{t-1}$ which greatly alleviates the calculation speed. The computations of gate states are shown as follows:
	\begin{equation}
	State_{i,t} = \sigma (W_{xi}x_{t}+W_{hi}h_{t-1}+b_{i})
	\end{equation}
	where $State_{i,t}$ denotes input gate state, hidden gate state and forget gate state at time $ t $. It is apparent that hidden state $h_{t-1}$ in the last time is used for updating all gate states at this time. 
	
	Different from the previous methods, we use the recurrent model proposed by \cite{lei2017sru} to capture temporal information. The advantages of using SRUs for modeling temporal dynamics are three folds: First, it would boost performance if substitute LSTM in some case. Second, SRUs operates faster than LSTM. Third, deep SRUs could be trained well by adding skip connection. 
	
	The SRUs architecture used in the proposed MRRN is defined as follows:
	
	\begin{align}
	\widetilde{x}_{t}&=Wx_{t}\\
	f_{t}&=\sigma (W_{f}x_{t}+b_{f})\\
	r_{t}&=\sigma (W_{r}x_{t}+b_{r})\\
	c_{t}&=f_{t}\odot c_{t-1}+(1-f_{t})\odot \widetilde{x}_{t}\label{four}\\
	h_{t}&=r_{t}\odot g(c_{t}) + (1-r_{t})\odot x_{t}\label{five}
	\end{align}
	where $f_{t}, r_{t}$ are sigmoid gates referred to the forget gate and reset gate, $g(\cdot)$ is the hyperbolic tangent function, as illustrate in fig.\ref{sru}. SRUs breaks the dependency by completely dropping $h_{t-1}$ in the recursion, which simplifies the state computation and discloses more parallelism while retaining the strong capability for representations. It is worth to mention that the update of internal states $c_{t}$ still depends on the previous states $c_{t-1}$. When input vectors $x_{t}$ are passed into SRUs module, $\widetilde{x}_{t}, f_{t}, R_{t}$ at each time-step are computed at once. Eq.\ref{four} and Eq.\ref{five} operations are elementwise.
	
	As mentioned above, we average the activation tensor:
	\begin{equation}
	x_{t}=\frac{1}{N}\sum_{i=1}^{N}A_{l,i}^{t}
	\end{equation}
	as a fixed-length feature vector of $frame_{t}$ and put it into SRUs at time-step $ t $ resulted in the representation $ r_{t} $. For fusing predictions at all time-steps, we employ both mean pooling and max pooling to obtain multiple types of video representations.
	
	Thus, our model consists of two phrases, see fig\ref{piplines}. In the CNN phrases, we encode N continual video frames belonged to one video as a feature sequence $X=(x_{1}, ..., x_{N})$, where $x_{t}\in R^{C}$ ($C\in[128, 256, 512]$) and in the recurrent phrase, the probability distribution over action categories is calculated by the following equations:
	\begin{align}
	P_{mean}(y=j)&=\frac{exp(\frac{1}{N}\sum_{t=1}^{N}W_{jt}r_{t})}  {\sum_{j=1}^{M}exp(\frac{1}{N}\sum_{t=1}^{N}W_{jt}r_{t})}\label{mean}\\
	P_{max}(y=j)&=\frac{exp(max\forall_{t\subseteq [1, N]} W_{jt}r_{t})}{\sum_{j=1}^{M}exp(max\forall _{t\subseteq [1, N]}W_{jt}r_{t})}\label{max}
	\end{align}
	where $W_{jt}$ represents the weight parameters mapping inputs of the recurrent model at time $t$ to action $j$, M is the number of hidden units. Eq.\ref{mean} and Eq.\ref{max} denote mean pooling prediction and max pooling prediction respectively. Finally, we use the following formula to combine different-level representations and make the final prediction. 
	\begin{equation}
	P_{final}=a\times P_{H}+b\times P_{M}+c\times P_{L}
	\end{equation}
	where $ P_{(\cdot) } $ in the left of the equation refers to the prediction of different-level models and $P_{final}$ is the final prediction produced by combined model. According to the performance of three different-level RRNs, we assign 0.7, 0.2 and 0.1 as a, b and c respectively in following experiments.

	\section{Experiments}
	In this section, we first introduce two popular challenging datasets, HMDB-51 dataset and UCF-101 dataset\cite{Soomro2012UCF101}. Then we specify implement details of all experiments involved in this paper.
	\subsection{Dataset}
	UCF-101 dataset and HMDB-51 dataset are challenging action recognition benchmarks because of limited data. UCF-101 dataset has 13320 videos that are collected from YouTuBe and organized as 101 action categories. Every kind of behavior was conducted by 25 humans and everyone did more than one times from 4 to 7. The action categories can be divided into five types: Human-Object Interaction, Body-Motion Only, Human-Human Interaction, Playing Musical Instruments and Sports.
	
	Moreover, the HMDB-51 dataset is collected from various sources, mostly from movies. This dataset contains 3570 training clips and 1530 testing clips belonged to 51 distinct categories. These actions can be roughly divided into five groups, general facial actions, facial actions with object manipulation, general body movements, body movements with object interaction and body movements for human interaction. We use HMDB-51 dataset to illustrate the general relations between hyper-parameter and performance. Besides, both datasets are used to find the better manner of pooling and explore the secret of different-level representations.
	
	\subsection{Implement details}
	\begin{figure*}[htbp]
		\centering
		\includegraphics[width=17cm]{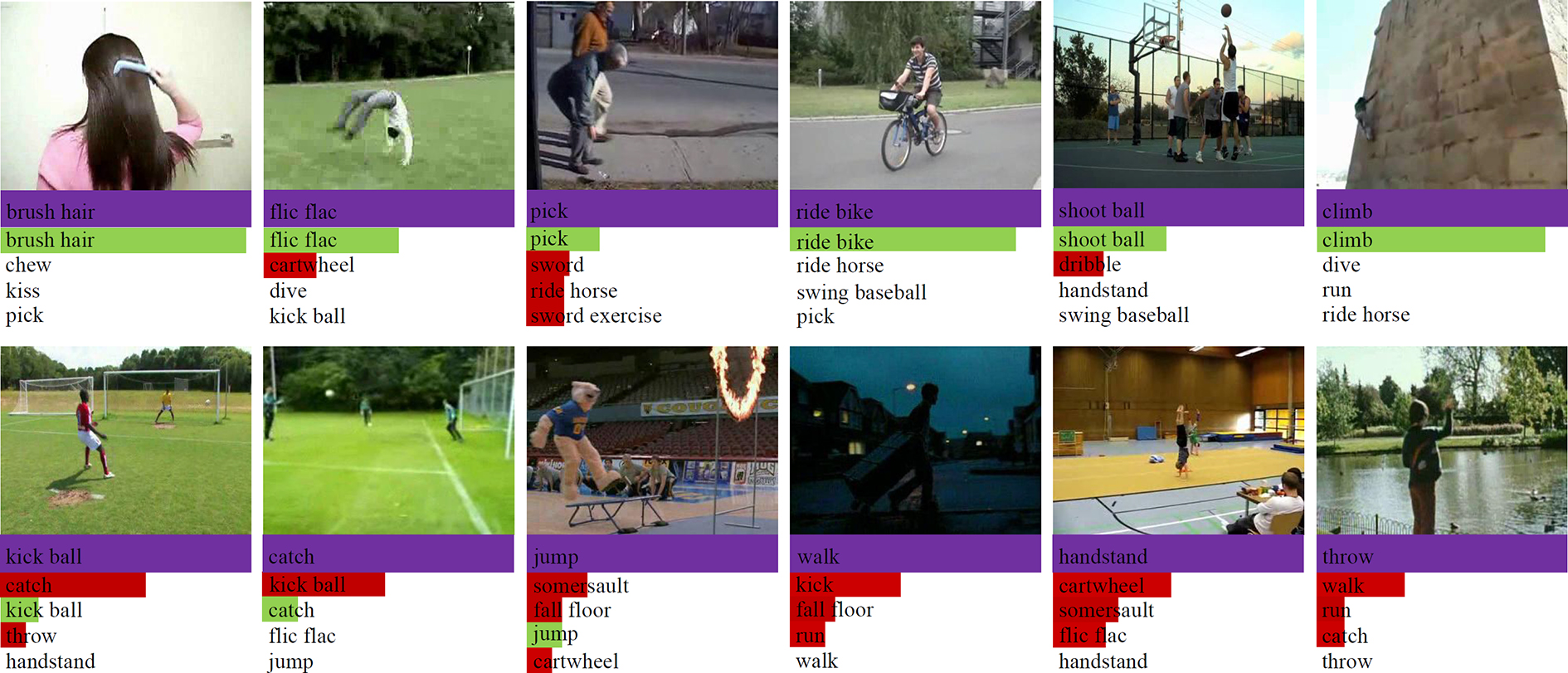}
		\caption{\textbf{Predictions on HMDB-51.} \small{Under each sub-picture, purple, green and red bars indicate the ground truth label, correct prediction, and incorrect prediction respectively sorted in decreasing confidence. The length of bars is on behalf of the probability of corresponding actions. The first row shows correct predictions and the second shows wrong predictions.}}
	\end{figure*}
	In the experiment of hyper-parameters, high-level ResNet-34 was used to extract frame representations resulted in 512-dimensional feature vectors. For investigating hyper-parameters, we experimented with the number of hidden units and set it to 256, 512 or 1024 units. Similarity, the number of layers was set to 3, 4 or 5 layers. While in the experiment of hierarchical features, the setting of input to SRUs is consistent with the shape of activation tensors produced by ResNets. For low-, mid-, high-level frame representations, the input size of the recurrent model is 128, 256 and 512 respectively. Nonlinear transformation in recurrent computations used sigmoid activation function and we added the hyperbolic tangent activate layers on the top of stacked SRUs for predictions at each time-step. Max pooling and mean pooling were performed over all time-steps. It is worth to mention that we did not adjust any structures of original ResNet-34 and retrain it to extract frame representations.
	
	The orthogonal weight initialization introduced by \cite{Saxe2013Exact} was used to initialize SRUs and all models were trained from scratch in an end to end scheme. Adam optimizer algorithm\cite{Kingma2014Adam} and CrossEntropyLoss function were used to optimize all models with mini-batch size 28 for 12 epochs over the entire datasets. The initial learning rate was set to 1e-5 for the first 8 epochs and changed to 1e-6 for the rest. We initially tried to set learning rate 1e-4 which led to the convergence quickly at first but obtained the relatively poor performance in the end. We adopt the dropout regularization ratio 0.5 for linear transformations to improve the generalization of models.
	
	The setting of our experiments followed the practice in \cite{DBLP:journals/corr/SharmaKS15}. The smaller side of the image was scaled to 256 and then a $224\times224$ region was randomly cropped from the rescaled image or its horizon flip with 50\% probability for data augmentation. Besides, the values of image pixel were transformed from $H\times W\times C$ in the range [0, 255] to the tensor of the shape $C\times H\times W$ in the range [0, 1.0] and we normalized each channel in the way of $channel=(channel-mean)/std$ with mean(0.485, 0.456, 0.406) and std(0.229, 0.224, 0.225). The same pre-process approaches were used in training and testing except that we used randomly crop from rescaled image in training while center crop in testing without horizon flip.
	
	Videos were split into 30 frames clips with a stride of 8 frames and the maximum of clips split from one video was set to 20. We would loop the video if necessary when the length of videos is insufficient to 30 frames. Thus, we obtained 21147 clips for training that each served as an individual training sample and 8791 clips for testing. In testing, we fused all predictions of clips belonged to one video by averaging their softmax scores.
	
	Depending on experiments in this paper, we select the best configuration and pooling manner as default settings to evaluate the effectiveness of MRRN. Here, we design three separate models according to low-, mid- and high-level representations and are combined by averaging their weighted softmax scores, as illustrated in fig.\ref{piplines}.
	
	All experiments were carried out on the first split of HMDB-51 dataset or UCF-101 dataset and performed on 4 NVIDIA Titan X GPUs based on the available public deep-learning framework Pytorch.

	\section{Results}
	\begin{figure*}[hthb]
		\centering
		\includegraphics[width=16cm]{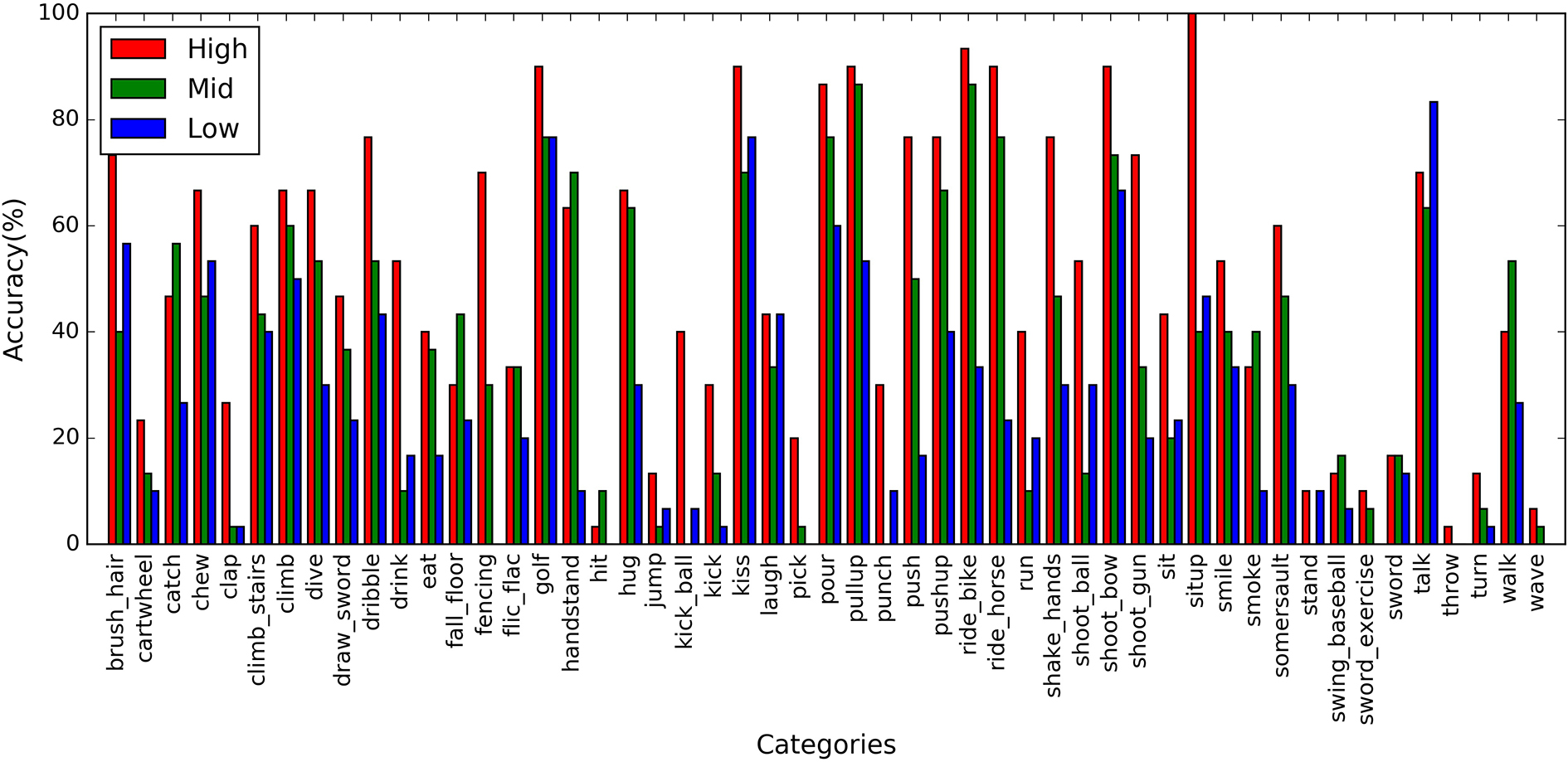}
		\caption{\textbf{Per-class results on the first split of HMDB-51.} This figure shows the per-class results predicted by different-level RRN. The red, green and blue bars represent the confidence of each class produced by low-, mid-, high-level RRN.}\label{bars}
	\end{figure*}
	In the following section, we experimentally demonstrated the effectiveness of MRRN on action recognition benchmarks and compared it with state-of-the-art models. This section contains four parts. In the first part, we studied networks with various hyper-parameters on HMDB-51 dataset. We leveraged different-level RRN with two pooling manners to verify the contribution to action recognition on both datasets in the second part. The third part is the critical evaluations of MRRN. Finally, we computed both time complexity and space complexity to verify the efficiency.
	\subsection{Hyper-Parameters}
	We first investigated the networks with 256, 512 or 1024 hidden units and 3, 4 or 5 layers, the mutual combinations between them. In addition, we chose the high-level ResNets as the default extractor so that output size of it is 512-dimensional feature vector. That means the input size of SRUs is fixed as 512 units.
	\renewcommand\arraystretch{1.1}
	\begin{table}[htbp]
		\centering
		\caption{Experiments on Hyper-parameters}\label{hyper_parameters}
		\begin{tabular}{cccc}
			\hline
			setting&3layers&4layers&5layers\\
			\hline
			256 hiddenunits&48.10&48.10&43.92\\
			512 hiddenunits&48.17&49.28&46.21\\
			1024 hiddenunits&50.78&49.61&47.71\\
			\hline
			
		\end{tabular}
	\end{table}

	Table \ref{hyper_parameters} reports various experimental accuracies under different settings. The results from Table \ref{hyper_parameters} illustrate the general tendency that from top right to bottom left, the testing accuracies are constantly rising. From the perspective of hidden units, we can find that the capability of distinguishing actions increases with the number of hidden units regardless of how many layers we set. There are some subtle differences in layers where no clear linear relations between layers and accuracies. We propose the hypothesis that there may be an inverse relation that the property would degrade with layers under certain conditions if we make abundant experiments. Therefore, we could try the setting of 3 layers with 1024 hidden units first for use of SRUs. Due to the fact that we imposed the restriction on the input size of SRUs, we did not take the influence of changed input vectors into account. Therefore, one thing we obliged to think about is the impact of changed inputs for setting hyper-parameters.
	
	In the Table \ref{hyper_parameters}, the SRUs with 3 layers and 1024 hidden units achieves the best performance by 50.78\%. The review of Recurrent Neural Networks\cite{Lipton2015A} reveals that the expressive power of hidden states grows exponentially with the number of nodes. Increasing the number of hidden units leads to a great improvement where the performance gap between maximum and minimum reaches to 6.9\%, so we fixed the number of hidden units as 1024 and layers as 3 in the remainder of this section, although we obtained the bad performance in experiments of low-level and mid-level representations. We will analyze it further in the next sub-section.

	\subsection{Hierarchy Features}
	This section aims to verify whether hierarchical features can benefit the performance. In addition, we also compared max pooling and mean pooling methods to understand the differences between them. We picked activation tensors produced by different-level ResNet-34 as targets. The results conducted on both datasets are listed in Table \ref{pooling}.
	\begin{table}[htbp]
		\centering
		\caption{Experiments on Hierarchy Features}\label{pooling}
		\begin{tabular}{ccccc}
			\hline
			Dataset&Pooling&Low&Mid&High\\
			\hline
			
			\multirow{2}{*}{HMDB-51}&mean&26.60&36.80&50.78\\
			&max&25.36&35.82&47.78\\
			\multirow{2}{*}{UCF-101}&mean&46.50&64.16&81.38\\
			&max&45.23&64.96&81.25\\
			\hline
		\end{tabular}
	\end{table}
	
	It is apparent that high-level RRN with mean temporal pooling obtains the supreme performance. In the case of low-level representations and mid-level representations, we changed the learning rate to 1e-4 for accelerating the convergence because the convergent speed of two shallower models with default learning rate is far slower than the deeper model. The performance gaps between different-level models are so large that shallower video representations cannot be used to classify action alone. The study\cite{Zagoruyko2016Paying} constructs several functions that map the 3D tensor to the 2D tensor along the channel dimension and discovers that different layers in the network focus on different parts of the image. Low-level RRN and mid-level RRN concentrate on redundant details and ignore full object, which leads to decline. Whereas from the fig.\ref{bars} we can see that the performance of high-level RRN is lower than the other two in particular classes. Consequently, it is beneficial to integrate different video representations as they are highly complementary to each other.
	
	In the comparison of time series pooling, mean pooling is superior to max pooling in most case. Our analyses draw the preliminary conclusion that max pooling is sensitive to the noise in the convolutional network because it takes the maximum from the given dimensions of the activation tensor, while mean pooling considers entire activation values. So, we assign mean pooling as default time series pooling, unless otherwise stated.
	\subsection{Evaluation of MRRN}
		\begin{figure*}[hthb]
	\centering
		\includegraphics[width=12cm]{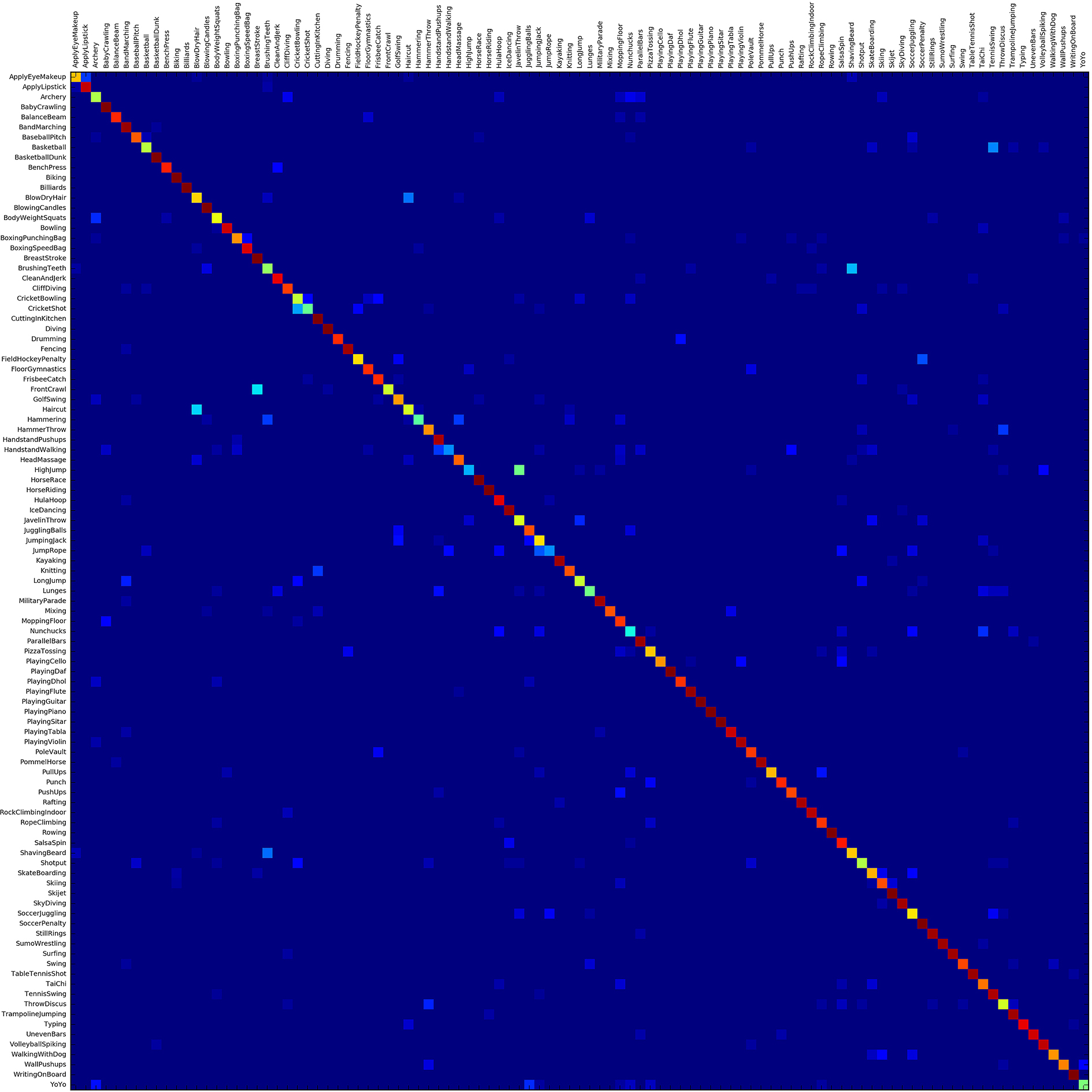}
		\caption{\textbf{Confusion matrix of MRRN model on the first split of UCF-101.} Each column represents the predicted class, and each row represents the ground truth class. The brightness of squares in diagonal represents the proportion of correct predictions. This figure shows results that the combined model makes predictions on the first split of UCF-101 and achieves 81.9\% accuracy.}\label{confusion}
	\end{figure*}
	\begin{table}[htbp]
	\centering
	\caption{Comparisons with State-of-the-art}\label{comparison}
	\begin{tabular}{ccc}
		\hline
		Model&HMDB-51&UCF-101\\
		\hline
		iDT+SVM\cite{INRIA2013LEAR, wang:2011:inria-00583818:1} & 52.1& 76.2\\
		iDT+HD encodings\cite{DBLP:journals/corr/PengWWQ14} &\textbf{61.1} & \textbf{87.9}\\
		\hline
		Slow Fusion Network\cite{Karpathy2014Large}&-&65.4\\
		LRCN\cite{DBLP:journals/corr/DonahueHGRVSD14} & - & 71.1\\
		Spatial ConvNet\cite{Simonyan2014Two} & 40.5 & 73.0\\
		Soft Attention\cite{DBLP:journals/corr/SharmaKS15} & 41.3 & -\\
		cLSTM\cite{Srivastava2015Unsupervised} & 44.1 & 75.8\\
		High RRN(Our Model) & 50.8 & 81.4\\
		MRRN(Our Model) & \textbf{51.3} & 81.9\\
		C3D Model\cite{Tran2014Learning} & - & \textbf{82.3}\\
		\hline
		Temporal ConvNet\cite{Simonyan2014Two} & - & 83.7\\
		scLSTM\cite{Ng2015Beyond} & 55.1 & 84.0\\
		cLSTM\cite{Srivastava2015Unsupervised} &-&84.3\\
		Two-Stream\cite{Simonyan2014Two} & 59.4 & 88.0\\
		Feature stacking\cite{Lan2014Beyond}& 65.4 & 89.1 \\
		TDD+iDT\cite{Wang2015Action}& 65.9 & 91.5\\
		RNN-FV+iDT\cite{Lev2015RNN}& \textbf{67.7} & \textbf{94.1}\\
		\hline
	\end{tabular}
	\end{table}
	Finally, we make the comparisons between our model and other competitive action recognition models on both UCF-101 dataset and HMDB-51 dataset. Table \ref{comparison} summarizes the results listed in original papers. We divide these comparisons into three sets. The upper set is simple features with linear SVM. The second set compares models that take only RGB data as input and the last combines multiple features to predict activities. 
	
	It is easy to spot that the deep learning models which use RGB image merely as input are inferior to the previous hand-crafted features based models though deep models can capture ample semantic information. One problem with this is the available training data is so limited that the deep learning models fail to learn the intrinsic trait of data. It is worth to mention that MRRN has not been pre-trained on any video datasets. In addition to this, our networks are with too many hyper-parameters that have the huge impact on the network performance and we have no clue to search for the best hyper-parameters but try. That is, we would be highly competitive with hand-crafted features based results if we give full play to the potential of MRRN.
	
	In contrast to the model in the second part, the MRRN boosts the performance to 51.3\% and 81.9\% on both datasets, which outperforms the majority of models and achieves the comparable result to the best accuracy. This indicates the combined model can merge information in different depth and there are some benefits to improve the precision, although the performance of two shallow models is relatively poor. We observe that our model outperforms 10.8\% and 8.9\% better than Spatial ConvNet\cite{Simonyan2014Two} on HMDB-51 dataset and UCF-101 dataset respectively which just learns semantic representations from image appearance, suggesting additional motion information is beneficial to action recognition and MRRN can well capture sufficient temporal dynamics in the video sequence. Compared with cLSTM, MRRN improves from 44.1\% to 51.3\% and 75.8\% to 81.9\% on both datasets, which shows MRRN learns more powerful spatiotemporal representations than cLSTM. 3D Convolutional network extracts spatiotemporal features from multiple adjacent frames and trains a linear SVM classifier, which is a little better than our model. It can be explained by the fact that multi-class linear SVM classifier has a stronger ability of the classification than softmax classifier in some case.
	
	Meanwhile, we also make the comparisons with models which use multiple features as input. When combined appearance information with optical flow information, the performance can be enhanced by a large margin. It is noted that the performance of cLSTM is lower than our model by 5.6\% when only using RGB data, but superior to ours by 2.4\% on UCF-101 dataset when combined with optical flow images. Despite all that, these methods have an obvious disadvantage of computation expensively compared to our model. 

	The confusion matrix for UCF-101 classification is shown fig.\ref{confusion} for displaying experimental results intuitively. Each column represents the predicted class, and each row represents the ground truth class. The higher brightness of squares in diagonal indicates the better prediction our model makes and vice versa. This figure shows the results that MRRN makes predictions on the first split of UCF-101 and achieves 81.9\% accuracy. 
	\subsection{Complexity Analysis}
	Finally, we calculated time complexity and space complexity of several competitive methods for verifying the efficiency. Time complexity indicates the computation complexity that estimates the time taken for training and inferring, while space complexity indicates the number of parameters which a model needs. Note that more training data is needed to train the model that has higher space complexity. To make a fair comparison with prior works, all experiments followed the original setting. Complexities are defined as follows:
	\begin{align}
	Time&\sim O(\sum_{l=1}^{D}M_{l}^{2}\cdot K_{l}^{2}\cdot C_{l-1} \cdot C_{l})\label{time}\\
	Space&\sim O(\sum_{l=1}^{D}K_{l}^{2}\cdot C_{l-1} \cdot C_{l})\label{space}
	\end{align}
	where $ D $ denotes the number of convolutional layers, $ M_{l}^{2} $ denotes the size of feature map at $ l $ layers, $ K_{l}^{2} $ denotes the size of kernel at $ l $ layers, $ C_{l-1} $ and $C_{l}$ denote the number of channels at $ l-1 $ layers and $ l $ layers respectively. 
		\begin{figure}[htbp]
		\centering
		\includegraphics[width=8cm]{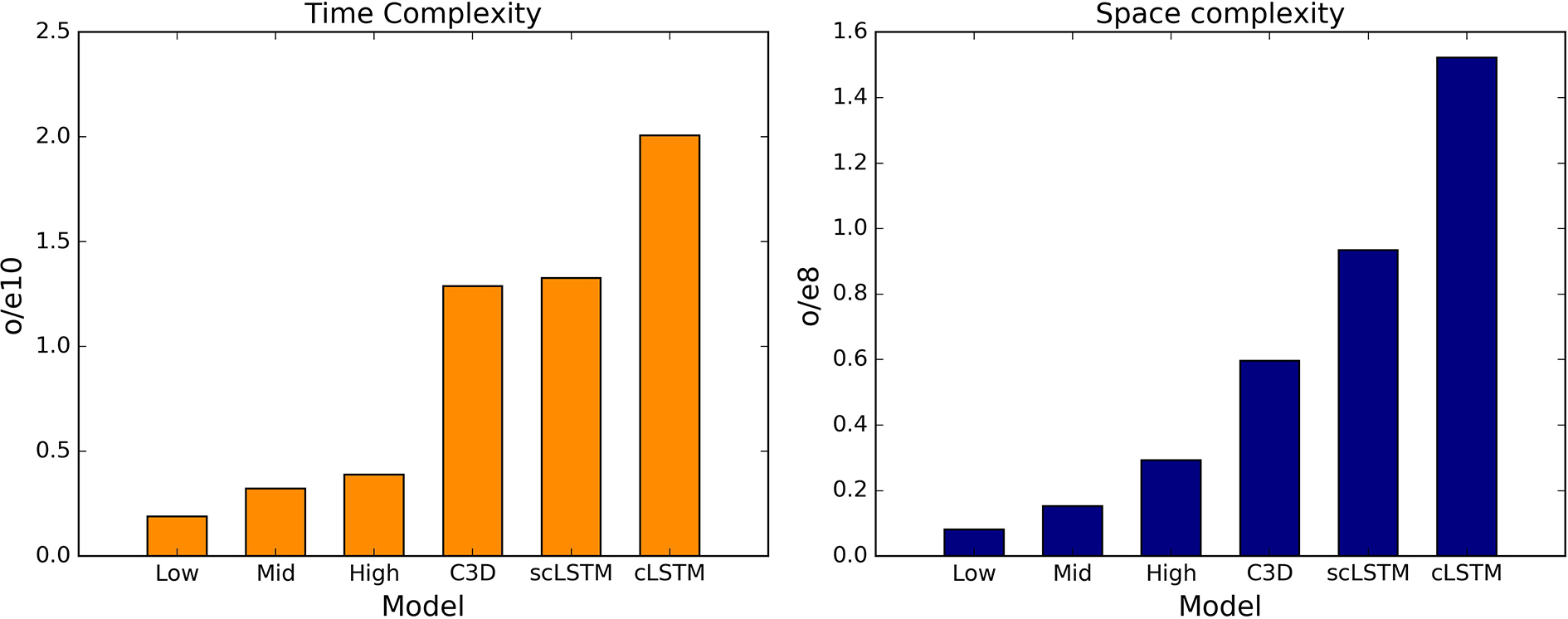}
		\caption{\textbf{Complexity.} \small{Our three sub-models are trained and tested in parallel, so we compute the complexity independently. The same process is applied to scLSTM and cLSTM. When computed complexities, we did not take bias into account for brevity. All experiments followed original settings.}}\label{complexity}
	\end{figure}

	The fig \ref{complexity} shows our three sub-models are lower than others in both time complexity and space complexity. C3D and scLSTM dramatically add complexity for abundant $3\times3$ convolutional kernels while cLSTM uses large-size feature maps in former layers for improving accuracies. This illustrates the trade-off between complexity and accuracies.

	\section{Conclusion}
	This work proposed a novel model, Multi-Level Recurrent Residual Networks(MRRN), which learns effective video representations for action recognition. We performed extensive evaluations on the model with various hyper-parameter settings, empirically illustrating the general tendency of performance. In addition, the combination of different-level models was shown that complementary representations further boost the accuracy. Moreover, our model has lower space complexity and time complexity by employing identity shortcut compared to the state-of-the-art.

	\section{Acknowledgements}

	This work was supported partly by the National Natural Science Foundation of China(61772067, 61472030, 61471032), the fundamental research funds for the central universities (2017JBZ108).

	\bibliographystyle{unsrt}

\end{document}